\documentclass[runningheads]{llncs}
\usepackage{graphicx}
\usepackage{comment}
\usepackage{amsmath,amssymb} % define this before the line numbering.
\usepackage{color}
\usepackage{multirow}
\usepackage{amssymb}% http://ctan.org/pkg/amssymb
\usepackage{pifont}% http://ctan.org/pkg/pifont
\usepackage{xcolor}

\usepackage{booktabs}
\usepackage{xspace}

\newcommand{\mymodel}{Commonality-Parsing Network\xspace}

\usepackage{hyperref}

\begin{document}
%\bibliographystyle{unsrt}
% \linenumbers
\pagestyle{headings}
\mainmatter
\def\ECCVSubNumber{596}  % Insert your submission number here

\title{\mymodel across Shape and Appearance for Partially Supervised  Instance Segmentation} % Replace with your title

%\title{Boundary-Aware Novel Objects Segmentation Network with Bi-directional Gate Mechanism and Position-sensitive Graph Convolutional Network}

\begin{comment}
\titlerunning{ECCV-20 submission ID \ECCVSubNumber} 
\authorrunning{ECCV-20 submission ID \ECCVSubNumber} 
\author{Anonymous ECCV submission}
\institute{Paper ID \ECCVSubNumber}
\end{comment}
%******************

%\begin{comment}
\titlerunning{\mymodel}
%\author{First Author\inst{1}\orcidID{0000-1111-2222-3333} \and
%Second Author\inst{2,3}\orcidID{1111-2222-3333-4444} \and
%Third Author\inst{3}\orcidID{2222--3333-4444-5555}}

\def\thefootnote{$*$}\footnotetext{Equal contribution}
\def\thefootnote{$\dagger$}\footnotetext{Corresponding author: Wenjie Pei}

\author{Qi Fan$^*$\inst{1} \and
Lei Ke$^*$\inst{1} \and
Wenjie Pei$^\dagger$\inst{2} \and
Chi-Keung Tang\inst{1} \and
Yu-Wing Tai\inst{1,3}}
\authorrunning{Q. Fan, L. Ke, W. Pei, C.K. Tang and Y.W. Tai}
\institute{Hong Kong University of Science and Technology \\
\email{\{qfanaa, lkeab, cktang, yuwing\}@cse.ust.hk} \and
Harbin Institute of Technology, Shenzhen \\
\email{\{wenjiecoder\}@gmail.com} \and
Kwai Inc.}
%\end{comment}
%******************
\maketitle

\begin{abstract}

Partially supervised instance segmentation aims to perform learning on limited mask-annotated categories of data thus eliminating expensive and exhaustive mask annotation. The learned models are expected to be generalizable to novel categories. Existing methods either learn a transfer function from detection to segmentation, or cluster shape priors for segmenting novel categories. We propose to learn the underlying class-agnostic commonalities that can be generalized from mask-annotated categories to novel categories. Specifically, we parse two types of commonalities: 1) shape commonalities which are learned by performing supervised learning on instance boundary prediction; and 2) appearance commonalities which are captured by modeling pairwise affinities among pixels of feature maps to optimize the separability between instance and the background. Incorporating both the shape and appearance commonalities, our model significantly outperforms the state-of-the-art methods on both partially supervised setting and few-shot setting for instance segmentation on COCO dataset. The code is available at {\color{magenta}\url{https://github.com/fanq15/CPMask}}.


\vspace{-3mm}
\end{abstract}

\section{Introduction}
\vspace{-2mm}
Instance segmentation is a fundamental research topic in computer vision due to its extensive applications ranging from object selection~\cite{li2004statistical}, image editing~\cite{rother2004grabcut,vezhnevets2005growcut} to scene understanding~\cite{li2003foreground}. Typical methods~\cite{chen2018masklab,he2017mask,huang2019mask,li2017fully,liu2018path} for instance segmentation have achieved remarkable progress, relying on the fully supervised learning on the precise mask-annotated data. However, this kind of pixel-level mask annotation is extremely labor-consuming and thus expensive to be performed on large amount of data which is typically required for deep learning methods. On the other hand, it is less expensive and more feasible to perform annotation of bounding box for instances, which motivates the newly proposed task: \emph{partially supervised instance segmentation}~\cite{hu2018learning,kuo2019shapemask}. It aims to learn instance segmentation models on limited mask-annotated categories of data, which can be generalized to new (novel) categories with only bounding-box annotations available. The partially supervised instance segmentation is much more challenging than the typical instance segmentation in full supervision. The major difficulty lies in how to learn the class-agnostic features for instance segmentation that can be generalized from the mask-annotated categories to novel categories. 

\begin{figure}[!t]
\centering
\includegraphics[width=0.85\linewidth]{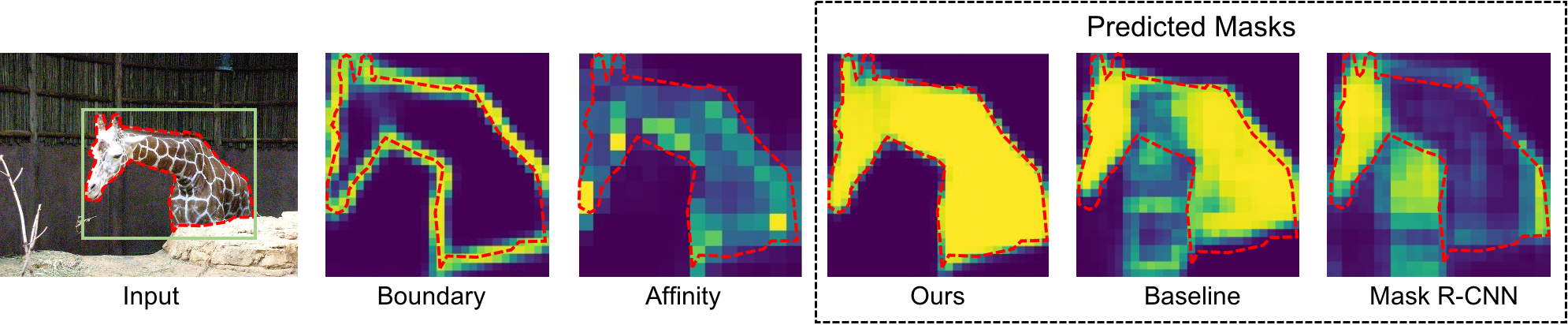}
\vspace{-0.15in}
\caption{Given an input image, our model captures shape commonalities by predicting instance boundaries and learns the appearance commonalities by modeling pairwise affinities among all pixels. The learned class-agnostic commonalities in both shape and appearance enable our model to segment more accurate mask than other models. Note that the similar background (wall color) misguides other methods. Herein, the affinity heatmap encodes the mean of the affinity maps for each instance pixel. The baseline model refers to basis framework of our model without commonality-parsing modules. The red dash line indicates the ground-truth of annotated mask.}
\label{fig:intro_example}
%\vspace{-0.3cm}
\vspace{-0.25in}
\end{figure}

A straightforward way for partially supervised instance segmentation is to directly extend existing fully supervised algorithms to segmentation of novel categories by class-agnostic training~\cite{pinheiro2015learning,pinheiro2016learning}, which treats all mask-annotated categories of instances involved in training as one foreground category and forces the model to learn to distinguish between foreground and background regions for segmentation. This brute-force way of class-agnostic training expects the model to learn all the generalized features between annotated and novel categories by itself, which is hardly achieved. As the initiator of the partially supervised instance segmentation, Mask$^X$ R-CNN~\cite{hu2018learning} transfers the visual information from the modeling of bounding box to the mask head through a parameterized transfer function. Subsequently, ShapeMask~\cite{kuo2019shapemask} seeks to extract the generic class-agnostic shape features across different categories by summarizing a collection of shape priors as reference for segmenting new categories. 

Whilst both Mask$^X$ R-CNN and ShapeMask have distinctly advanced the performance of partially supervised instance segmentation, there are two important features have not been fully exploited. First, the generalized \textbf{appearance} features that shared across different categories, e.g., similar hairy body surface between dogs and cats or similar textures on the furniture surface, are not explicitly explored. These class-agnostic appearance features can be potentially generalized from mask-annotated categories of data to novel categories for segmentation. Second, the common \textbf{shape} features that can be generalized across different categories are not explicitly learned in a supervised way, though ShapeMask refines the shape priors by simply clustering the annotated masks and adapts them to a given novel object. In this work we intend to tackle the partially supervised instance segmentation by fully exploiting these two features. %addressing these two issues. 

We propose to capture the underlying commonalities which can be generalized across different categories by supervised learning for partially supervised instance segmentation. In particular, we aim to learn two types of generalized commonalities: 1) the shape commonalities that can be generalized between different categories like similar instance contour or similar instance boundary features; 2) the appearance commonalities that shared among categories of instances owning similar appearance features such as similar texture or similar color distribution. 
%To capture the shape commonalities, we focus on learning discriminative features to predict boundaries between instances (foreground) and the background. The supervised learning of boundaries prediction is performed in a class-agnostic manner to capture the generalized shape features. To learn the appearance commonalities, we propose to model the affinities among pixels of feature maps in such a supervised way that the pixels belonging to a instance (foreground region) have closer affinities than the affinities between foreground and background pixels. Since the supervision on modeling affinities is performed for all involved categories by the same parameterized module, the learned appearance commonalities are expected to be able to generalized across different categories.  
The resulting model, \mymodel (denoted as CPMask), can be trained in an end-to-end manner. Consider the example in Fig.~\ref{fig:intro_example}, to segment the giraffe in the red bounding box, our model extracts its shape information by predicting the boundaries of giraffe and captures the appearance information by modeling the pairwise affinities among pixels. Taking into account both the shape and appearance information, our model is able to predict more accurate segmentation mask than other models. It is worth noting that although giraffe is a novel category whose mask-annotation is not provided in the training data, our model is able to accurately predict its boundary and affinity due to the learned class-agnostic commonalities w.r.t. both shape and appearance information. 

We evaluate our model on two settings on COCO dataset: 1) partially-supervised instance segmentation, in which partial categories are provided with the ground-truth for both bounding boxes and segmentation masks while the other (novel) categories are only provided with the annotated bounding boxes during training; 2) few-shot instance segmentation, in which each of the novel categories only contain a small number of training samples (with both annotated bounding boxes and masks). Our model outperforms the state-of-the-art performance significantly on both settings. We further qualitatively demonstrate the generalization ability of our model by directly applying our trained model on COCO dataset to other 9 datasets with various scenes.
It is worth mentioning that our model is more effective given fewer mask-annotated categories of training data compared to methods for fully supervised (routine) instance segmentation.
%Additionally, the experiments on routine fully-supervised instance segmentation shows that our model (based on single-stage backbone) also performs best among single-stage methods, which indicates the learned shape and appearance commonalities are also beneficial for fully-supervised setting. 
To conclude, our contributions includes:
\vspace{-5pt}
\begin{itemize}
    \item We design a supervised learning mechanism for predicting instance boundaries to learn the class-agnostic shape commonalities that can be generalized from mask-annotated categories to novel categories. 
    \item We propose to model the affinities among pixels of feature maps in a supervised way to optimize the separability between the instance region and the background and learn the class-agnostic appearance commonalities that can be generalized to novel objects. %in such a supervised way that the pixels belonging to a instance (foreground region) have closer affinities than the affinities between foreground and background pixels. Thus we are able to optimize the separability between the instance region and the background and learn the class-agnostic appearance commonalities that can be generalized to novel objects. 
    \item Incorporating both learned shape and appearance commonalities, our model substantially outperforms state-of-the-art methods on COCO dataset for instance segmentation in both partially supervised and few-shot setting. %, demonstrated by extensive experiments on COCO dataset. Besides, our model (based on single-stage backbone) also achieves best performance among all single-stage instance segmentation methods in fully supervised instance segmentation.
\end{itemize}

\vspace{-4mm}

\section{Related Works}
\vspace{-2mm}
%Instance segmentation is a fundamental yet challenging task, which not only provides pixel-level object category classification, but also distinguishes individual objects as separate entities.
%and assigns different labels to different instances although they belong to the same category.

\noindent{\bf Conventional Instance Segmentation.} is fully supervised by numerous high-quality pixel-level annotations~\cite{dai2015convolutional,dai2016instance,girshick2014rich,hariharan2014simultaneous,hariharan2015hypercolumns,hayder2017boundary,pinheiro2015learning,pinheiro2016learning}.
Lots of methods have made great progress on this task by embracing the classical ``detect then segment" paradigm, which first generates detection results using the powerful two-stage detector and then segments each object in the bounding box. 
Mask R-CNN~\cite{he2017mask} attaches one simple mask predictor on Faster R-CNN~\cite{ren2015faster} to segment each object in the box.
PANet~\cite{liu2018path} merges multi-level features to enhance the performance. 
FCIS~\cite{li2017fully} and MaskLab~\cite{chen2018masklab} use position-sensitive score maps to encode the segmentation information. Kong and Fowlkes~\cite{Kong_2018_CVPR} propose to use pairwise pixel affinity for instance segmentation.
Mask Scoring R-CNN~\cite{huang2019mask} introduces a mask IoU branch to predict the mask quality and then selects good mask results accordingly. HTC~\cite{chen2019hybrid} fully leverages the relationship between detection and segmentation to build a successful instance segmentation cascade network. Most recently, some works attempt to build instance segmentation network on the one-stage detector~\cite{lin2017focal,tian2019fcos} for its simplicity and efficiency. 
In YOLACT~\cite{bolya2019yolact}, a set of prototype masks and coefficients are used to assemble masks for each instance. CenterMask~\cite{lee2019centermask} builds an attention-based mask branch on FCOS~\cite{tian2019fcos} for fast mask prediction. Compared to these previous works, our model mainly targets for novel objects segmentation by learning shape and appearance commonalities, although it also achieves superior performance in the fully supervised task.

\noindent{\bf Instance Segmentation for Novel Objects.}
Generalizing instance segmentation model to novel categories with limited annotations is meaningful and challenging, which mainly has three different settings:  
{\bf Weakly supervised} instance segmentation methods are developed to use weak labels to segment novel categories where the training samples are only annotated with bounding boxes~\cite{khoreva2017simple,remez2018learning} or image-level labels~\cite{ahn2019weakly,zhou2018weakly} without pixel-level annotations. {\bf Few-shot supervised} instance segmentation~\cite{yan2019metarcnn} is proposed to solve this problem by imitating the human visual systems to learn new visual concepts with only a few well-annotated samples. 
{\bf Partially supervised} instance segmentation is formulated in a mixture of strongly and weakly annotated scenario where only a small subset of base categories are well-annotated with both box and mask annotations while the novel categories only have box annotations. 
In Mask$^X$ R-CNN~\cite{hu2018learning}, a parameterized weight transfer function is designed to transfer the visual information from detection to segmentation while ShapeMask~\cite{kuo2019shapemask} learns the intermediate concept of object shape as the prior knowledge. 
Different from the above two works, which solve the partially supervised segmentation task either from transfer learning perspective or utilizing additional shape priors, our model focuses on learning class-agnostic features with great generalization ability by parsing the shape and appearance commonalities and clearly outperforms the existing methods by a large margin.
%Although our method is targeted at the partially supervised instance segmentation, it can be directly applied to the more challenging few-shot supervised setting for novel object segmentation.

%\noindent{\bf Boundary information and Non-Local} Boundary information has been used in semantic segmentation~\cite{bertasius2015high,chen2016semantic,yu2018learning,ding2019boundary} and salient object detection~\cite{zhao2019egnet,qin2019basnet,luo2017non,wang2019salient} tasks served as an important cue to improve performance. However, the boundary cues are rarely tapped into the problem of instance segmentation. We find that boundary is an important general cue to help instance segmentation network to generalize to novel objects, and we try to introduce instance segmentation system the under-explored boundary information to endow it generalization ability. {\bf Non-local} network is first proposed by~\cite{wang2018non} to capture long-range dependencies between any two positions. It has been applied in the backbone to obtain powerful relation-aware features~\cite{pmlr-v97-zhang19f,hu2019local,chen2019graph}. Different from these works, we improve the non-local network to capture the appearance commonality shared among categories and thus can be easily generalized to novel objects.

\vspace{-3mm}

\section{\mymodel}
\vspace{-2mm}
The crux of performing novel instance segmentation is to learn the underlying commonalities that can be generalized from the mask-annotated categories to novel categories. To surmount this crux, our \mymodel performs class-agnostic learning for partially supervised instance segmentation by two proposed modules: 1) Boundary-Parsing Module for learning shape commonalities and 2) Non-local Affinity-Parsing Module for learning appearance commonalities. We will first present the overall framework of the proposed \mymodel, then we will elaborate on the aforementioned two modules specifically designed for class-agnostic learning.
\begin{figure}[t]
\centering
\includegraphics[width=0.85\linewidth]{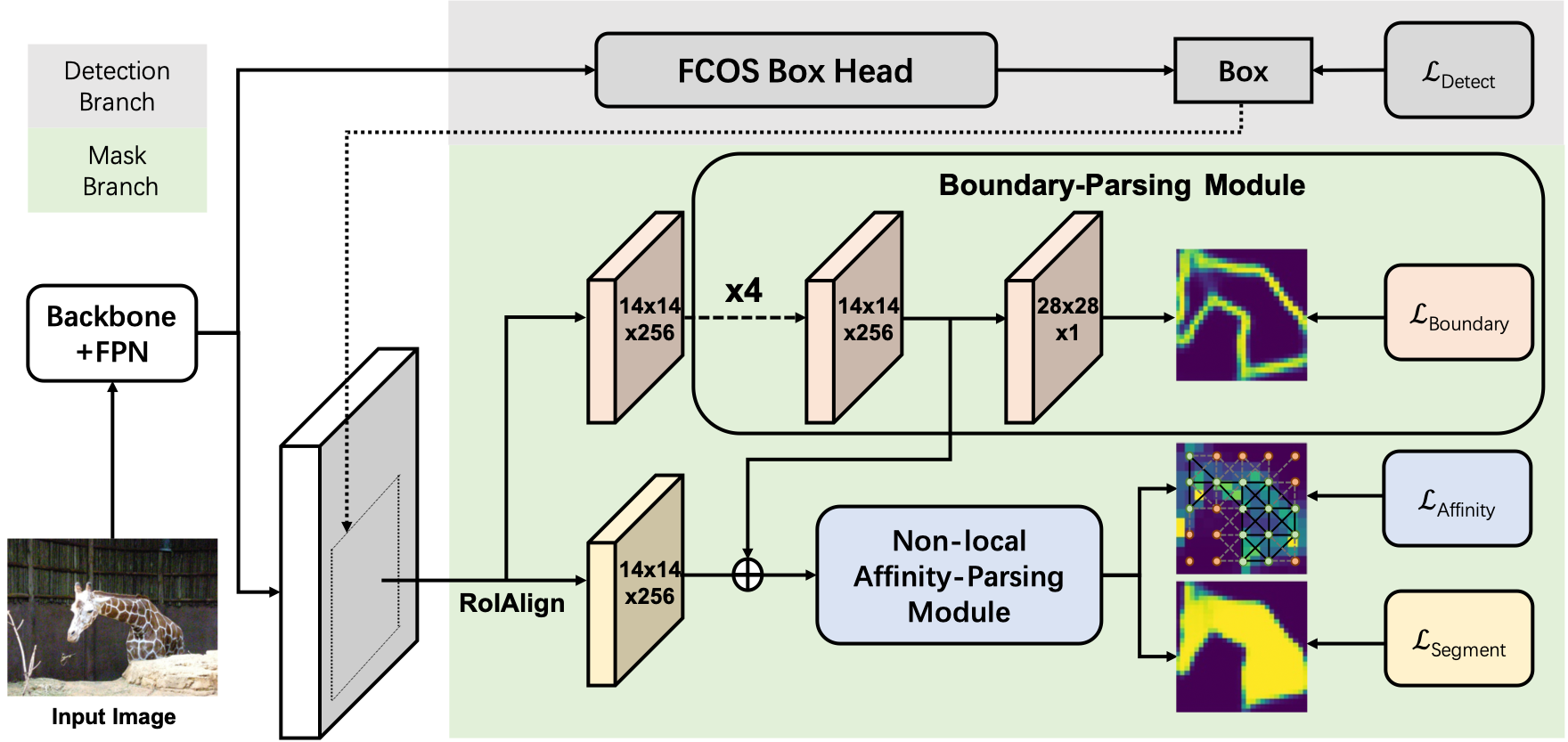}
\vspace{-0.15in}
\caption{Architecture of our \mymodel which consists of a detection branch and a mask branch. The cropped RoI feature based on predicted bounding boxes is first processed by Boundary-Parsing Module of the mask branch for predicting instance boundaries to guide the learning of shape commonalities in intermediate feature maps. Then the feature maps are fed into Non-local Affinity-Parsing Module (presented in Fig.~\ref{fig:affinity}) to learn the appearance commonalities by modeling the pairwise affinities among all pixels in feature maps. Finally, the feature maps incorporating both shape and appearance commonalities are used for mask prediction.}
\label{fig:network}
%\vspace{-0.3cm}
\vspace{-0.2in}
\end{figure}

\vspace{-2mm}
\subsection{Class-Agnostic Learning Framework}
\vspace{-1mm}
Fig.~\ref{fig:network} presents the architecture of our \mymodel. Following typical models~\cite{chen2018masklab,he2017mask,liu2018path} for instance segmentation, our model contains two branches: 1) the object detection branch in charge of predicting bounding boxes as instance proposals, and 2) the mask branch for predicting segmented masks for the instance proposals obtained from the object detection branch. 

We adopt FCOS~\cite{tian2019fcos}, which is an excellent one-stage detection model, as our object detection backbone. %Note that it is readily replaceable by any other object detection frameworks~\cite{lin2017focal,liu2016ssd,redmon2017yolo9000}. 
As illustrated in Fig.~\ref{fig:network}, a backbone network equipped with FPN~\cite{lin2017feature} is first employed to extract intermediate convolutional features for downstream processing. The object detection branch is then utilized to predict bounding boxes with positions as well as categories for potential instances. In the training phrase, supervision on both the position prediction and the category classification is performed to guide the optimization of the backbone network and FPN as in~\cite{tian2019fcos}:
\begin{small}
\vspace{-2mm}
\begin{equation}
    \mathcal{L}_{\text{Detect}} = \mathcal{L}_{\text{regression}} + \mathcal{L}_{\text{centerness}} + \mathcal{L}_{\text{classification}}.
\end{equation}
\end{small}
\vspace{-0.25in}

The mask branch is responsible for segmenting each of target instances predicted by the object detection branch. It is composed of two core modules designed specifically for class-agnostic learning by parsing the commonalities across both the shape and appearance features: Boundary-Parsing Module and Non-local Affinity-Parsing Module. These two modules are trained on a small set of mask-annotated categories of data (termed as base categories) and the learned inter-category commonality of both shape and appearance information enables our model to perform instance segmentation on novel categories of image data.  

\vspace{-2mm}
\subsection{Boundary-Parsing Module for Learning Shape Commonality}
\vspace{-1mm}
Boundary-Parsing Module is designed to learn the underlying commonalities with respect to the shape information that can be generalized from the mask-annotated categories to mask-unseen novel categories of data. Specifically, the Boundary-Parsing Module focuses on learning to predict the boundaries between the instance (foreground) and the background. The rationale behind this design is that there are common shape features shared among different categories on discrimination of the instance-background boundaries, which can be leveraged during class-agnostic learning for instance segmentation of novel categories. Besides, accurate boundary localization is able to explicitly contribute to the mask prediction for segmentation, which has been proved by many works~\cite{arbelaez2009contours,arbelaez2010contour,bertasius2015high,chen2016semantic,ding2019boundary,luo2017non,qin2019basnet,wang2019salient,yu2018learning,zhao2019egnet}.
Hence, we perform supervised learning for the prediction of boundaries to learn the shape commonalities among different categories.  

There are several ways to design the structure of Boundary-Parsing Module and we just investigate a straightforward yet effective way: four $3\times 3$ convolutional layers with ReLU as the activation functions, followed by one upsampling layer and one $1\times 1$ convolutional layer to output one channel of feature map as boundary predictions. The Boundary-Parsing Module is trained with the boundary loss:

\vspace{-0.1in}
\begin{small}
\begin{equation}
\vspace{-3mm}
    \mathcal{L}_{\text{boundary}} = \mathcal{L}_{\text{BCE}}(\mathcal{F}_{B}(\mathbf{X}), \mathcal{GT}_{B}),
\end{equation}
\end{small}

\vspace{-2mm}\noindent
where $\mathcal{L}_{\text{BCE}}$ denotes the binary cross-entropy loss, $\mathcal{F}_{B}$ denotes the nonlinear transformation functions by Boundary-Parsing Module, $\mathbf{X}$ is the RoI feature cropped by the \emph{RoIAlign} operation corresponding to a target instance predicted by the object detection branch and $\mathcal{GT}_{B}$ is the off-the-shelf boundary ground-truth that can be readily obtained from mask annotations.

\vspace{-0.1in}
\subsection{Non-local Affinity-Parsing Module for Learning Appearance Commonality}
\label{sec:affinity}
Similar categories tend to share similar appearance commonality, e.g., similar hairy body surface between dogs and cats, or similar texture on the furniture surface.
%(\peicomment{any better examples? similar steel material on the machines?}). 
This kind of appearance commonalities can be leveraged for class-agnostic learning to generalize the instance segmentation to novel categories. Therefore, we propose Non-local Affinity-Parsing Module to learn the appearance commonalities across different categories by parsing the affinities among pixels of feature maps in a non-local way. The pixels belonging to an instance (in the foreground region) are expected to have much closer affinities than the affinities between foreground and background pixels.

\begin{figure}[t]
\centering
\includegraphics[width=0.85\linewidth]{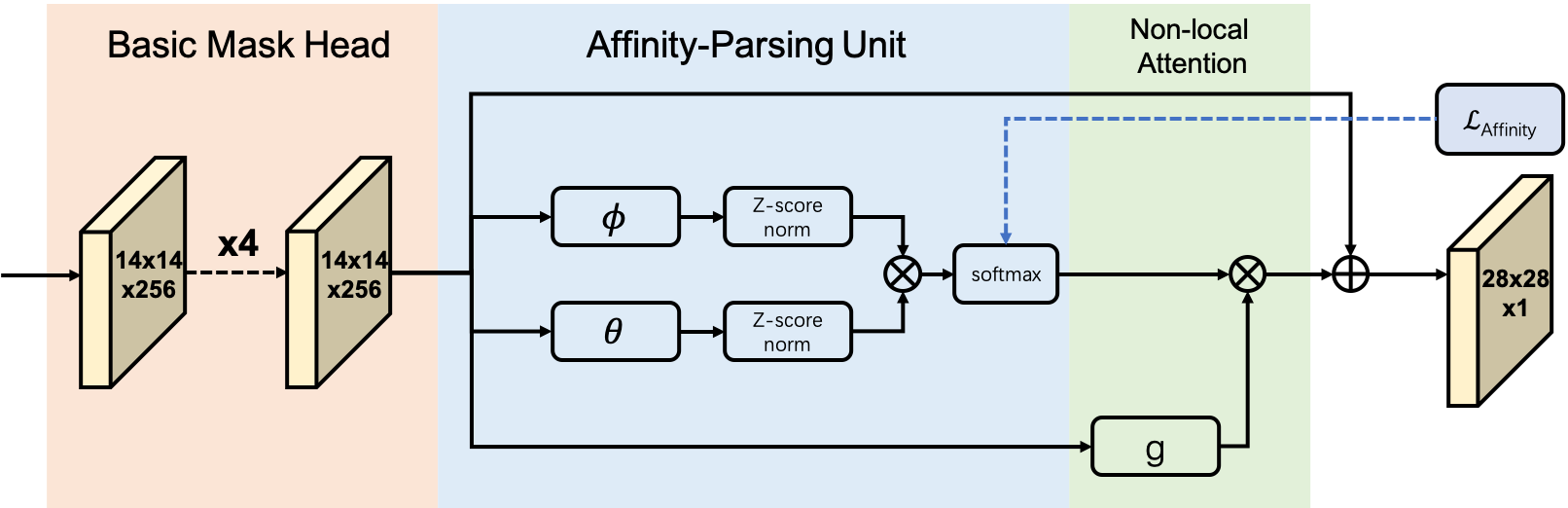}
\vspace{-0.15in}
\caption{Architecture of our Non-local Affinity-Parsing Module, which is composed of three units. The Basic Mask Head processes the input feature with four convolutional layers with $3 \times 3$ kernel and ReLU. Subsequently, the Affinity-Parsing unit performs supervised learning to model the pairwise affinities among pixels in feature maps. Finally, the non-local attention is employed to coordinate feature maps based on affinities to enable our model perceive more context information and increase the appearance separation between the instance and the background. Herein, ``$\otimes$" denotes the matrix multiplication and ``$\oplus$" represents element-wise addition.}
\label{fig:affinity}
\vspace{-0.25in}
%\vspace{-3mm}
\end{figure}

Formally, given the RoI feature $\mathbf{X}$ after \emph{RoIAlign} operation for an instance proposal, we first fuse it with the output feature maps $\mathcal{F}_B(\mathbf{X})$ from Boundary-Parsing Module by a simple attention module which incorporates the shape commonality information by weighted element-wise additions. Then the nonlinear transformation $\mathcal{G}$ by four convolutional layers is performed on the fused features as a basic mask head of operations:
\begin{small}
\begin{equation}
    \mathbf{C} = \mathcal{G}(\mathbf{X} \oplus \mathcal{F}_B(\mathbf{X}))).
\end{equation}
\end{small}

\vspace{-3mm}\noindent
The obtained feature maps $\mathbf{C} \in\mathrm{R}^{c\times h \times w}$, with $c$ feature maps of size $h\times w$, is then fed into the non-local affinity-parsing unit for modeling affinity. 
Specifically, we model the affinity between the pixel at ($i, j$) and the pixel at ($m, n$) in a latent embedding space by:
\begin{small}
\begin{align}
\vspace{-3mm}
\begin{split}
    & \mathbf{A}_{(<i,j>, <m,n>)} = f\Big [\frac{(\theta(\mathbf{C}_{i,j})-\mu_{i,j})}{\sigma_{i,j}}, \frac{(\phi(\mathbf{C}_{m,n})-\mu_{m,n})}{\sigma_{m,n}}
    \Big ],
%    & \mathbf{A} = \text{softmax} (\mathbf{A}),
    \end{split}
\end{align}
\end{small}

\vspace{-4mm}\noindent
where $\mathbf{C}_{i,j}\in \mathrm{R}^c$ corresponds to the vectorial representation (in channel dimension) for the pixel at $(i,j)$ and the same goes for $\mathbf{C}_{m,n}$. Herein, $\theta, \phi$ are embedding functions and $f$ is a kernel function for encoding affinity. In practice, we opt for the dot-product operator for $f$, which is a typical way of modeling similarity. $\mu$ and $\sigma$ are the mean value and the standard deviation respectively. Note that here we apply the \emph{z-score} normalization for both $\theta(\mathbf{C}_{i,j})$ and $\phi(\mathbf{C}_{m,n})$ to ease the convergence during optimization. 
%To constrain the value of affinity $\mathbf{A}$ to lie in $[0,1]$, \emph{softmax} operator is employed. 

Larger affinity value indicates closer relationship while smaller affinity value implies larger difference. We expect that the affinities between pixels belonging to an instance (foreground) region are much higher than that between foreground and background pixels. To this end, we introduce a supervision signal to guide the optimization to achieve the desired affinity distribution. In particular, we impose an affinity constraint to maximize the affinities among pixels in the foreground region $Fg$ and minimize the affinities between foreground $Fg$ and background $Bg$ pixels:
\begin{small}
\begin{align}
\vspace{-5mm}
\begin{split}
& \mathbf{A} = \text{softmax} (\mathbf{A}),\\
& \mathcal{L}_{\text{Affinity}}= \mathcal{L}_{1}(1, \hspace{-5pt} \sum_{\begin{tiny}\begin{subarray}{c}<i,j>\in Fg\\ <m,n>\in Fg\end{subarray} \end{tiny}} \hspace{-10pt} \mathbf{A}_{<i,j>, <m,n>}) + \mathcal{L}_{1}(0, \hspace{-5pt} \sum_{\begin{tiny}\begin{subarray}{c}<i,j>\in Fg\\ <m,n>\in Bg\end{subarray}\end{tiny}} \hspace{-10pt} \mathbf{A}_{<i,j>, <m,n>}).
\end{split}
\end{align}
\end{small}

\vspace{-2mm}\noindent
Here we first normalize $\mathbf{A}$ using a \emph{softmax} operator and then impose the loss function that encourages the sum of affinities among foreground pixels to be close to 1 for more appearance affinities while pushing the affinities between foreground and background pixels to be 0 for larger appearance separation.

The supervised learning on the affinity distribution enables our model to perceive the appearance separability between the foreground (instance) and background regions. To further increase this appearance separation, we propose to coordinate feature maps by explicitly incorporating the learned affinities in a non-local attention manner~\cite{wang2018non,pmlr-v97-zhang19f}:
\begin{small}
\begin{equation}
\vspace{-2mm}
    \widetilde{\mathbf{C}}_{i,j} = \sum_{\forall <m,n>} \mathbf{A}_{<i,j>,<m,n>} \cdot g(\mathbf{C}_{m,n}),
\end{equation}
\end{small}

\vspace{-1mm}\noindent
where $g$ is a embedding function. Here we coordinate the vectorial representation for the pixel at $(i,j)$ in the feature maps by attending each pixel with the corresponding  affinity. Such coordination on feature maps enables our model to perceive the context of whole image region with affinity-based attention, thus resulting in more separation of appearance between foreground and background and closer affinities among pixels in foreground (instance) region, which is beneficial for learning appearance commonalities and instance segmentation. 

Together with original feature maps $\mathbf{C}$, the output coordinated feature maps $\Tilde{\mathbf{C}}$ from the Non-local Affinity-Parsing Module is subsequently fed into one upsampling layer and one $1\times 1$ convolutional layer for the final prediction of segmented mask:
\begin{small}
\begin{equation}
\vspace{-2mm}
    \mathcal{L}_{\text{Segment}} = \mathcal{L}_{\text{BCE}}(\mathcal{F}_{1\times 1 conv}(\Tilde{\mathbf{C}} \oplus \mathbf{C}), \mathcal{GT}_{S}),
\end{equation}
\end{small}

\vspace{-2mm}\noindent
where $\mathcal{F}_{1\times 1 conv}$ denotes the nonlinear transformation functions by $1\times 1$ convolutional layer and $\mathcal{GT}_{S}$ is the ground-truth mask annotations.

\vspace{-0.1in}
\subsection{End-to-End Parameter Learning}
The whole model of our \mymodel can be trained in an end-to-end manner on two different types of training data:
\vspace{-1mm}
\begin{itemize}
    \item For the mask-annotated training data in base categories, the model is optimized
by integrating all the aforementioned loss functions:
\begin{small}
\begin{equation}
\vspace{-2mm}
    \mathcal{L} = \lambda_1\mathcal{L}_{\text{Detect}} + \lambda_2\mathcal{L}_{\text{Boundary}} + \lambda_3\mathcal{L}_{\text{Affinity}} + \lambda_4\mathcal{L}_{\text{Segment}},
\end{equation}
\end{small}

\vspace{-3mm}\noindent
where $\lambda_1$, $\lambda_2$, $\lambda_3,$ and $\lambda_4$ are hyper-parameter weights to balance the loss functions. In our implementation, they are tuned to be $\{1, 0.5, 0.5, 1\}$ respectively on a validation set. 
\item For the training data without mask-annotation in novel categories, we train the model with only detection loss, i.e., only the parameters in backbone network, FPN and detection branch are optimized:
\begin{small}
\begin{equation}
\vspace{-2mm}
    \mathcal{L} = \mathcal{L}_{\text{Detect}}.
\end{equation}
\end{small}
\end{itemize}

\vspace{-5mm}

\section{Experiments}
\vspace{-0.05in}

We conduct experiments on MS COCO dataset~\cite{lin2014microsoft} to evaluate our model. We first perform ablation study to investigate the effect of Boundary-Parsing Module and Non-local Affinity-Parsing Module, then we compare our model with state-of-the-art methods in three different settings for instance segmentation: 1) partially supervised setting, 2) few-shot setting and 3) fully supervised setting. 

\vspace{-0.15in}

\subsection{Experimental Setup}
\vspace{-0.05in}
\subsubsection{Evaluation Protocol.}
We follow the typical data split on COCO in our experiment: \emph{train2017} for training and \emph{val2017} for test. %Both of them contain 80 categories of samples. 
 In both of our experiments on partially supervised setting and few-shot setting, we split the 80 COCO categories into ``\emph{voc}" and ``\emph{non-voc}" category subsets where the \emph{voc} categories are those in PASCAL VOC~\cite{everingham2010pascal} dataset while the remaining categories are included in the \emph{non-voc} categories. Each time we select classes in one subset as base categories with annotations of both bounding boxes and masks, and those in the other subset as novel categories. Note that the training samples of novel categories have only bounding box annotation (no mask annotation) for partially supervised setting. For few-shot setting, each novel category in the training data only contains a small amount of samples with annotations of both bounding boxes and masks. %For all of our experiments, w
 	 We adopt the typical evaluation metrics for instance segmentation in our experiments, i.e., \emph{AP}, \emph{AP$_{50}$}, \emph{AP$_{75}$}, \emph{AP$_{S}$}, \emph{AP$_{M}$} and \emph{AP$_{L}$}.

\vspace{-0.2in}

\subsubsection{Implementation Details.}
SGD with Momentum is employed for training our model, starting with 1 K constant warm-up iterations. The batch size is set to 16 and initial learning rate is set to 0.01.  For efficiency, ResNet-50~\cite{he2016deep} is used as backbone network for ablation study and the input images are resized in such a way that the short side and long side are no more than 600 and 1000 pixels respectively (denoted as (600, 1000)). For other experiments on comparison with other methods, ResNet-101~\cite{he2016deep} backbone with multi-scale training is employed. %as the backbone network and the input images are resized to (1000, 1333) in the similar way as before.

\vspace{-0.15in}
\subsection{Ablation Study}
\vspace{-2mm}

 \setlength{\tabcolsep}{5.5pt}
 \begin{table}[!t]
 	\begin{center}{\small
			\resizebox{0.7\linewidth}{!}{
 			\begin{tabular}{l|cccccc} %|cccccc}
 				\hline\hline
 				&& \multicolumn{4}{c}{voc $\rightarrow$ non-voc} & \\ %&&\multicolumn{2}{c}{voc} & \\
 				\qquad\qquad model & $AP$ & $AP_{50}$ & $AP_{75}$ & $AP_{S}$ & $AP_{M}$ & $AP_{L}$ \\ %& $AP$ & $AP_{50}$ & $AP_{75}$  & $AP_{S}$  & $AP_{M}$ & $AP_{L}$ \\
 				%\noalign{\smallskip}
 				\hline
  				Baseline         		& 20.7 & 37.9 & 20.4 & 10.6 & 24.7 & 27.3 \\ 
 				Baseline + BM w/o FF    & 21.6 & 38.8 & 21.1 & 11.6 & 26.5 & 28.8 \\ 
 				Baseline + BM         	& 27.4 & 45.1 & 28.7 & 12.4 & 32.3 & 39.5 \\ 
 				Baseline + AM w/o AL    & 26.9 & 45.0 & 27.8 & 11.6 & 31.3 & 39.2 \\ 
 				Baseline + AM 	        & 27.2 & 45.2 & 28.3 & 11.7 & 31.5 & 40.3 \\ 
 				\hline
  				Baseline + BM + AM   	& 28.8 & 46.1 & 30.6 & 12.4 & 33.1 & 43.4 \\ 
 				\hline
 		\end{tabular}} }
 	\end{center}
 	\vspace{-0.1in}
 	\caption{Experimental results of ablation studies on the COCO {\it val} set. The models are trained on the \emph{voc} base categories and evaluated on the \emph{non-voc} novel categories. The ``BM" denotes the Boundary-Parsing Module, the ``AM" denotes the Non-local Affinity-Parsing Module, the ``FF" denotes fusing boundary feature to the mask head and the ``AL" denotes the affinity loss.}
 	%The ``Boundary" denotes our Boundary-Parsing Module, and the ``Affinity" denotes Non-local Affinity-Parsing Module. The first row corresponds to the baseline model.} %, %by the realistic protocol, 
 	%where good supports are used in fine-tuning or testing.}
 	\label{table:combination}
 	\vspace{-0.35in}
 \end{table}

\begin{figure}[!t]
\centering
\includegraphics[width=0.85\linewidth]{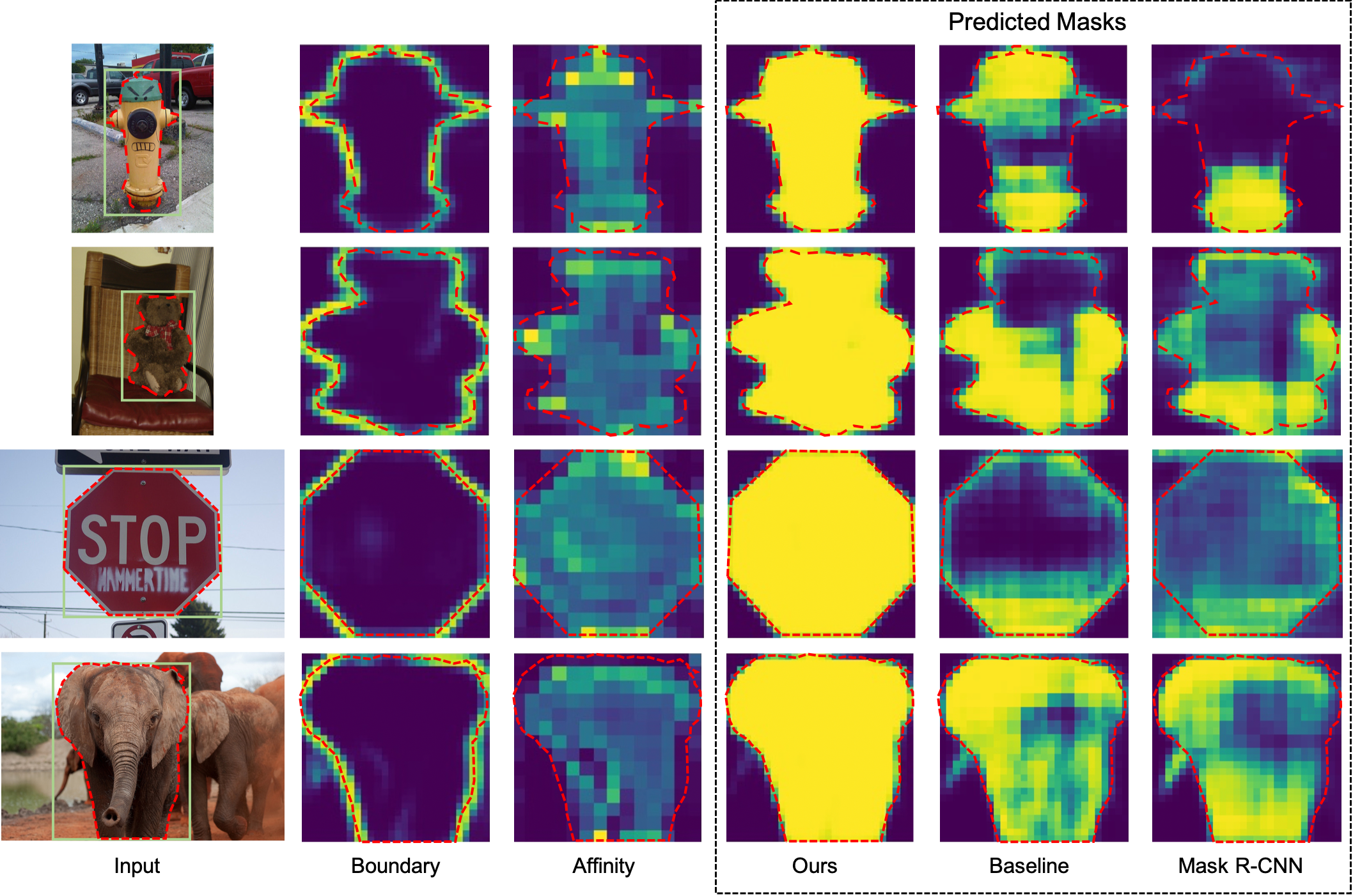}
\vspace{-0.15in}
\caption{Visualization of boundary heatmaps and affinity heatmaps learned by our model for four novel categories of cases. The red dash lines indicate the ground-truth mask. The affinity heatmap is obtained by calculating the mean of the affinity maps for each instance pixel. The {\color{yellow} yellow} color in the heatmap indicates higher response value and the {\color{blue} blue} color indicates lower response value.}
\label{fig:ablation}
\vspace{-0.25in}
%\vspace{-0.4cm}
\end{figure}

We investigate the effectiveness of our Boundary-Parsing Module and Non-local Affinity-Parsing Module by carrying out ablation experiments for partially supervised instance segmentation in this section.
The \emph{voc} classes is used as base categories and the \emph{non-voc} as novel categories.  
We refer to the variant of our model without Boundary-Parsing Module and Non-local Affinity-Parsing Module as \emph{Baseline} model. 
The class-agnostic version of Mask R-CNN~\cite{hu2018learning} is compared for reference in this section. 

%We conduct experiments which begin with the baseline model and then incrementally augment the model with Boundary-Parsing Module and the Non-local Affinity 
\smallskip\noindent\textbf{Quantitative Evaluation.} Table~\ref{table:combination} presents the experimental results. 
The baseline model obtains 20.7 AP on the novel categories. Boundary-Parsing Module improves the performance by 6.7 AP and explicitly adopting the boundary feature to guide the mask prediction is crucial for the overall performance. Non-local Affinity-Parsing Module promotes the performance by 6.2 AP and the better pixel relationship introduced by the affinity loss further boosts the performance to 27.2 AP.
%6.5 AP respectively, which indicates the effectiveness of both modules. 
Both the shape and appearance commonalities learned by these two modules from the base categories generalize well to the novel categories. %Furthermore, they also improve the base subset performance slightly, meaning that the learned commonality through these modules are also beneficial for the base categories. 
After integrating both modules, our model  achieves  28.8 AP which is distinctly better than the performance by each individual module. It implies that the learned shape and appearance commonalities contribute in their own way for instance segmentation.

\smallskip\noindent\textbf{Qualitative Evaluation.} To further reveal the mechanism of these two modules, we visualize boundary and affinity heatmaps on novel categories in Fig.~\ref{fig:ablation}. The affinity heatmap is obtained by calculating the mean of the affinity maps for each instance pixel in the Non-local Affinity-Parsing Module. %And the boundary heatmap is the one channel output feature map from the Boundary-Parsing Module. 
We observe that our model is able to accurately estimate instance boundaries to capture the shape commonalities. Meanwhile, the affinities between instance pixels are evidently higher (closer) than affinities between instance and background pixels, which indicates that appearance commonalities are well learned via affinity modeling for these novel categories. 
%The boundary heatmaps show that the instance shape is obviously captured by the instance boundary. And in the affinity heatmaps, only instance pixels have high affinity responses because the instance pixels have closer relationship with similar appearance commonality. 
Both of the shape and appearance commonalities can help our model to segment novel instances from  background, because the commonalities learned from these modules are successfully generalized from base categories to novel categories. By contrast, the baseline model without these two modules and the Mask R-CNN for fully supervised instance segmentation performs quite poorly on these cases.

\begin{figure}[!t]
\centering
\includegraphics[width=0.88\linewidth]{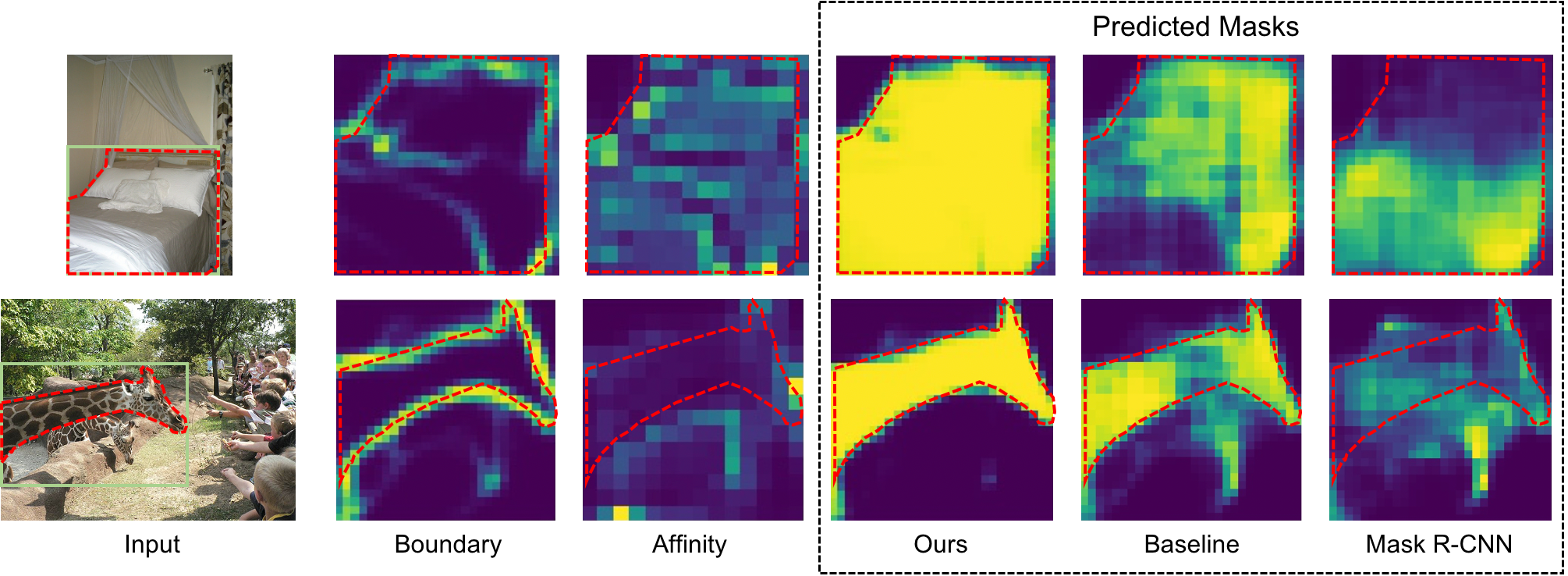}
\vspace{-0.1in}
\caption{Two Challenging examples that indicate the complementation between Boundary-Parsing Module and Non-local Affinity-Parsing Module.}
\label{fig:ablation_challenge}
\vspace{-0.1in}
%\vspace{-0.4cm}
\end{figure}

\begin{figure}[!t]
	\centering
	\includegraphics[width=0.78\linewidth]{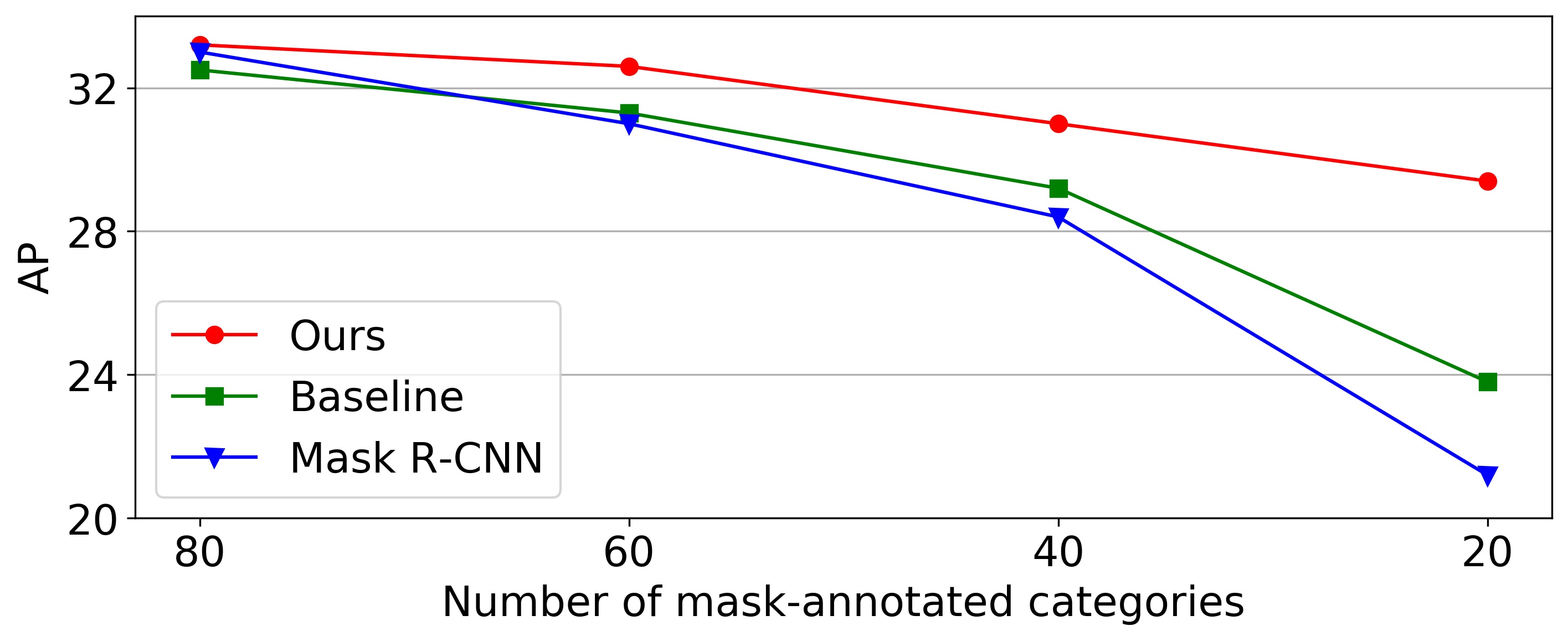}
	\vspace{-0.2in}
	\caption{The segmentation performance of different models on a fixed set of novel categories as a function of number of mask-annotated (base) categories. The novel categories are randomly selected from COCO dataset.}
	\label{fig:class_num}
	\vspace{-0.2in}
\end{figure}

\smallskip\noindent\textbf{Complementary advantages of Boundary (Shape) and Affinity (Appearance).} We present two challenging examples in Fig.~\ref{fig:ablation_challenge}. Our model cannot precisely estimate the boundary for the bed due to the confusing color differences. On the other hand, our Non-local Affinity-Parsing Module can tackle this problem very well based on appearance commonalities, which lead to the accurate mask prediction for bed. On the other hand, the affinity heat map is not precise for the giraffe due to the similar appearance of another overlapping giraffe behind. In such scenario with multiple instances of same category within a box, our Boundary-Parsing Module can still predict the boundaries very accurately.     

\smallskip\noindent\textbf{Evaluation of Generalization.} To further evaluate the ability of generalization from base (mask-annotated) categories to novel categories for our model,  we conduct experiments to investigate the effect of varying the number of mask-annotated categories in Fig.~\ref{fig:class_num}. The performances of the both baseline model and Mask R-CNN decay much faster than our model as the number of base categories for training decreases, which indicates that our method is particularly more effective given fewer annotated categories of training data compared to fully supervised methods and benefits from the class-agnostic learning of our model by Boundary-Parsing Module and Non-local Affinity-Parsing Module.

\vspace{-0.15in}
\subsection{Partially Supervised Instance Segmentation}
\vspace{-2mm}

In this section we compare our model to other state-of-the-art methods for partially supervised instance segmentation.

Table~\ref{table:partially} presents the quantitative results on COCO dataset with two sets of experiments: use \emph{voc} or \emph{non-voc} classes as the base categories and treat the remaining classes as novel categories.
Our model outperforms the state-of-the-art ShapeMask by a large margin: 3.8 AP on the \emph{non-voc} novel categories and 3.5 AP on the \emph{voc} novel categories respectively. Even compared to its stronger version equipped with NAS-FPN~\cite{ghiasi2019fpn} backbone which boosts the performance of both detection and segmentation, our model still performs better than ShapeMask. Besides, we also provide the \emph{oracle} performance which corresponds to the 
performance under full supervision and can be considered as the performance upper bound for partially supervised learning.  
We observe that the performance gap between our model and its oracle version is narrowed to 3.6/6.1 AP compared to 4.8/7.6 (4.4/7.4) AP by ShapeMask (ShapeMask with NAS-FPN) and 10.6/9.6 AP by Mask$^X$ R-CNN, indicating the advantages of agnostic learning by our specifically designed modules. 
% Note that our model even obtains competitive performance with the oracle models of ShapeMask and Mask$^X$ R-CNN under the same ResNet-101 backbone.
 %which demonstrates the powerful generalization ability to novel categories by our model. 
\setlength{\tabcolsep}{1.0pt}
\begin{table}[!t]
	\begin{center}{\small
			\resizebox{0.99\linewidth}{!}{
				\begin{tabular}{c|cccccc|cccccc}
					\hline\hline
					&& \multicolumn{4}{c}{voc $\rightarrow$ non-voc} &&& \multicolumn{4}{c}{non-voc $\rightarrow$ voc} & \\
					method & $AP$ & $AP_{50}$ & $AP_{75}$ & $AP_{S}$ & $AP_{M}$ & $AP_{L}$  & $AP$  & $AP_{50}$ & $AP_{75}$  & $AP_{S}$  & $AP_{M}$ & $AP_{L}$ \\
					\hline
					Mask R-CNN~\cite{he2017mask}            & 18.5 & 34.8 & 18.1 & 11.3 & 23.4 & 21.7 & 24.7 & 43.5 & 24.9 & 11.4 & 25.7 & 35.1 \\
					%Our Mask R-CNN        &  &  &           &          &  &  &  &  &           &          &  &  \\
					Mask GrabCut~\cite{hu2018learning}    & 19.7 & 39.7 & 17.0 & 6.4  & 21.2 & 35.8 & 19.6 & 46.1 & 14.3 & 5.1  & 16.0 & 32.4 \\
					Mask$^X$ R-CNN~\cite{hu2018learning}        & 23.8 & 42.9 & 23.5 & 12.7 & 28.1 & 33.5 & 29.5 & 52.4 & 29.7 & 13.4 & 30.2 & 41.0 \\
					ShapeMask~\cite{kuo2019shapemask}             & 30.2 & 49.3 & 31.5 & 16.1 & 38.2 & 38.4 & 33.3 & 56.9 & 34.3 & 17.1 & 38.1 & 45.4 \\
					ShapeMask (NAS-FPN)~\cite{kuo2019shapemask}   & 33.2 & 53.1 & 35.0 & 18.3 & {\bf 40.2} & 43.3 & 35.7 & 60.3 & 36.6 & {\bf 18.3} & {\bf 40.5} & 47.3 \\
					CPMask (Ours)                  & {\bf 34.0} & {\bf 53.7} & {\bf 36.5} & {\bf 18.5} & 38.9 & {\bf 47.4} & {\bf 36.8} & {\bf 60.5} & {\bf 38.6} & 17.6 & 37.1 & {\bf 51.5} \\
					\hline
					Oracle Mask$^X$ R-CNN~\cite{hu2018learning}     & {\color{gray}34.4} & {\color{gray}55.2} & {\color{gray}36.3} & {\color{gray}15.5} & {\color{gray}39.0} & {\color{gray}52.6} & {\color{gray}39.1} & {\color{gray}64.5} & {\color{gray}41.4} & {\color{gray}16.3} & {\color{gray}38.1} & {\color{gray}55.1} \\
					Oracle ShapeMask~\cite{kuo2019shapemask}      & {\color{gray}35.0} & {\color{gray}53.9} & {\color{gray}37.5} & {\color{gray}17.3} & {\color{gray}41.0} & {\color{gray}49.0} & {\color{gray}40.9} & {\color{gray}65.1} & {\color{gray}43.4} & {\color{gray}18.5} & {\color{gray}41.9} & {\color{gray}56.6} \\
					Oracle ShapeMask (NAS-FPN)~\cite{kuo2019shapemask}      & {\color{gray}37.6} & {\color{gray}57.7} & {\color{gray}40.2} & {\color{gray}20.1} & {\color{gray}44.4} & {\color{gray}51.1} & {\color{gray}43.1} & {\color{gray}67.9} & {\color{gray}45.8} & {\color{gray}20.1} & {\color{gray}44.3} & {\color{gray}57.8} \\
					Oracle CPMask (Ours)   & {\color{gray}37.6} & {\color{gray}58.2} & {\color{gray}40.2} & {\color{gray}19.9} & {\color{gray}42.6} & {\color{gray}54.2} & {\color{gray}42.9} & {\color{gray}67.6} & {\color{gray}46.6}          &    {\color{gray}21.6}      & {\color{gray}42.1} & {\color{gray}58.9} \\
					\hline
		\end{tabular}}}
	\end{center}
	\vspace{-0.1in}
	\caption{Experimental results of partially supervised instance segmentation on the COCO {\it val} set. The ``voc $\rightarrow$ non-voc" means that we use the \emph{voc} classes as base categories and the \emph{non-voc} as novel categories, and vice versa. The oracle models indicates the upper-bound performance for reference which are trained on masks from all categories (in full supervision).} %, %by the realistic protocol, 
	%where good supports are used in fine-tuning or testing.}
	\label{table:partially}
	\vspace{-0.23in}
\end{table}

 \begin{figure}[!t]
 	\centering
 	\includegraphics[width=0.87\linewidth]{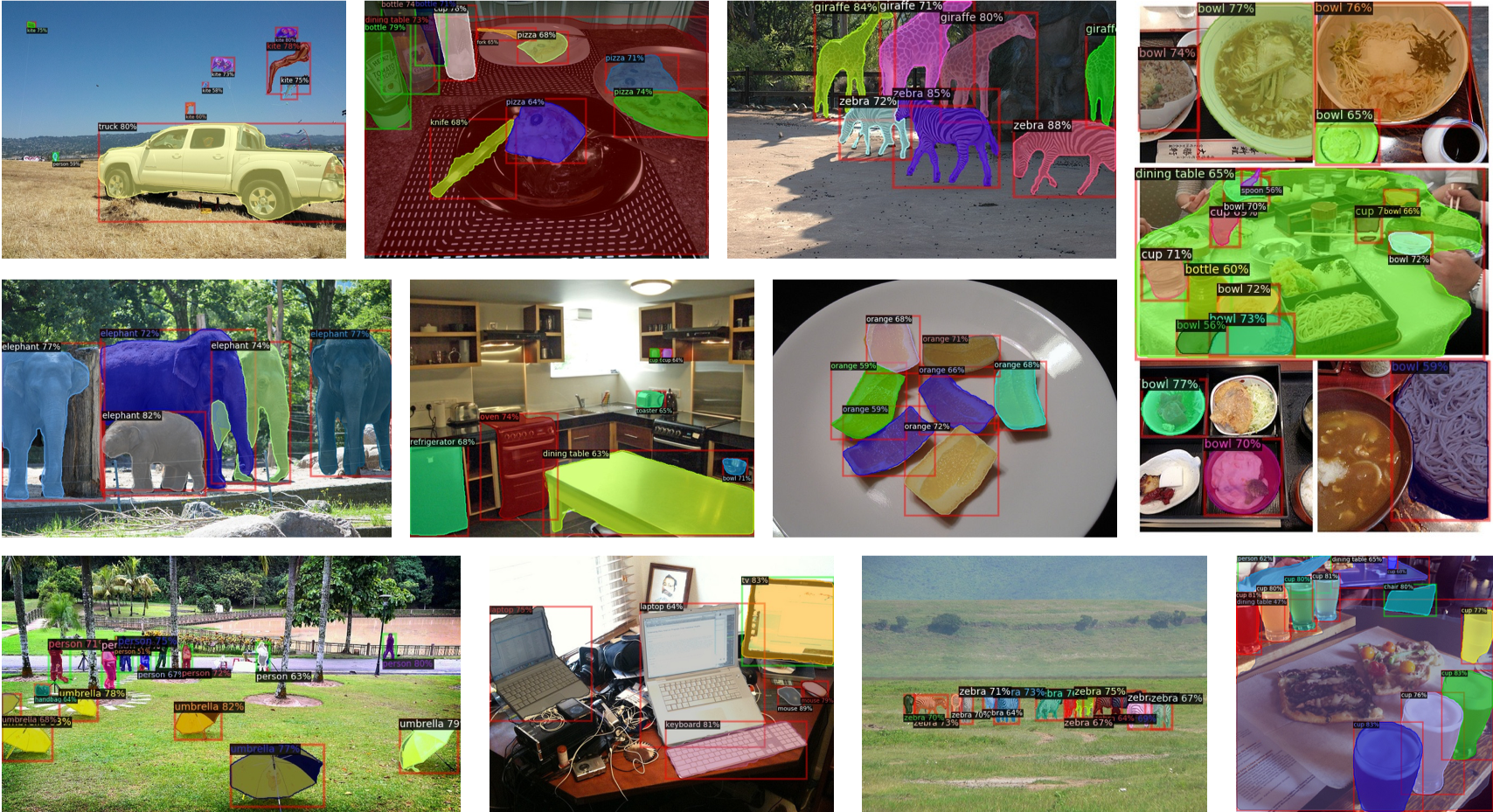}
 	\vspace{-0.1in}
 	\caption{Qualitative results on novel COCO categories. We use \emph{voc} classes as the base (mask-annotated) categories for training. }
 	\label{fig:coco}
 	\vspace{-0.3in}
 \end{figure}

\begin{figure}[!t]
	\centering
	\includegraphics[width=0.86\linewidth]{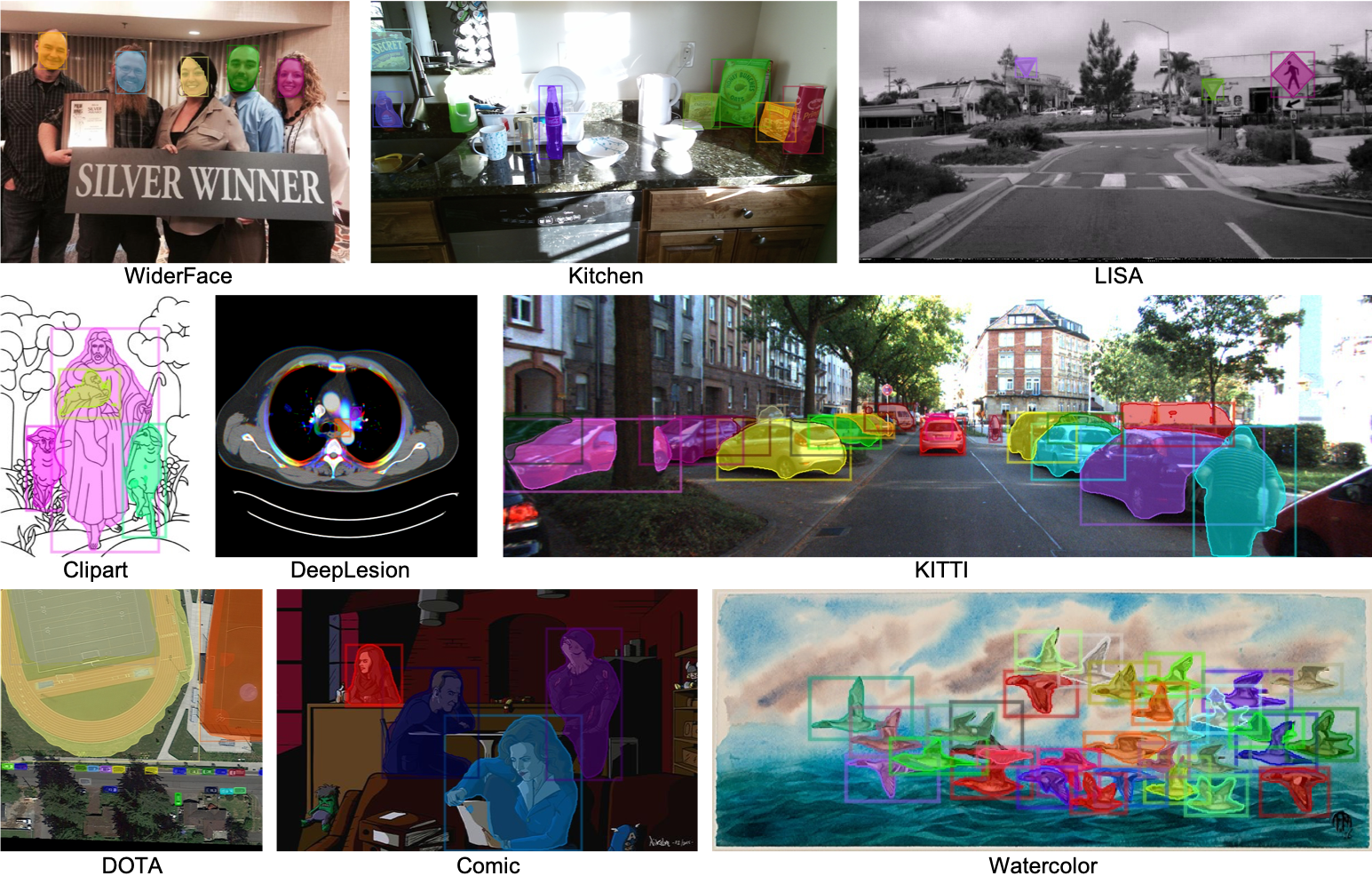}
	\vspace{-0.2in}
	\caption{Qualitative results of generalization by our model to 9 different datasets. The model is only trained on COCO and directly applied on these datasets.}
	\label{fig:applications}
	\vspace{-0.2in}
\end{figure}

 Fig.~\ref{fig:coco} shows qualitative results on multiple samples that randomly selected from COCO dataset including various scenes, which shows that our model is able to segment all different kinds of objects precisely, even for quite small ones. 
\noindent\textbf{Application on other datasets.} We further qualitatively demonstrate our model on other 9 datasets across various styles and domains~\cite{wang2019towards}:  Clipart~\cite{inoue2018cross}, Comic~\cite{inoue2018cross}, Watercolor~\cite{inoue2018cross},  DeepLesions~\cite{yan2018deep}, DOTA~\cite{xia2018dota},  KITTI~\cite{geiger2012we}, LISA~\cite{mogelmose2012vision}, Kitchen~\cite{georgakis2016multiview}, and WiderFace~\cite{yang2016wider}. %Clipart, Comic and Watercolor are in watercolor, clipart and comic styles respectively; DeepLesion is a medical CT image dataset about lesions; DOTA is an aerial image dataset; KITTI and LISA are collected from moving vehicles; Kitchen collects common kitchen objects; WiderFace is a human face dataset. 
	It is worth noticing that this is a much harder task due to the cross-dataset generalization. Specifically, we train our model on COCO dataset and feed it ground-truth boxes to obtain the segmentation results on these datasets. As shown in Fig.~\ref{fig:applications}, our model successfully segments novel objects from various domains. %generalizes well to novel categories in various domains.

\vspace{-3mm}
\subsection{Few-shot Instance Segmentation}
\vspace{-2mm}
In this section, we directly apply our model to the challenging few-shot instance segmentation without any network adaption.
Few-shot instance segmentation is another challenging task for novel categories. In this task, the model is first trained on base categories with numerous training samples and then generalizes to novel categories with only a few (10 or 20 shots) training samples by direct fine-tuning. Following Meta R-CNN~\cite{yan2019metarcnn}, the \emph{non-voc} classes is used as base categories with full samples per category and the \emph{voc} as the novel categories with only 10/20 training samples per category. For fair comparison, we follow Meta R-CNN~\cite{yan2019metarcnn} and use ResNet-50 as backbone and input image size is resized to (600, 1000). Note that the annotations of both bounding box and mask are provided for training samples in novel categories in the few-shot setting. 

As shown in Table~\ref{table:few-shot}, our model outperforms Meta R-CNN (the state-of-the-art method) by 2.7/3.9 AP in the 10/20-shot settings. Even equipped with the Faster R-CNN detector like Meta R-CNN, our model still performs much better. Although not specifically designed for few-shot learning, our model still obtains the state-of-the-art performance, demonstrating that our proposed model is not limited to the partially supervised learning, and is general for other novel instance segmentation tasks. %To validate the effectiveness of our model in the extreme annotation scenarios, we further propose a more challenging setting:  ``no base", in which the model is only trained on the novel categories with limited training samples, i.e., no pre-training on base categories.  
%Table~\ref{table:few-shot} shows that our model is still comparable with or better than other methods in the ``no base" setting. These results in the few-shot setting manifest the generalization and effectiveness of our model in the extreme annotation scenarios.

\setlength{\tabcolsep}{1pt}
\begin{table}[!t]
	\begin{center}{\small
				\resizebox{0.8\linewidth}{!}{
			\begin{tabular}{c|cccccc|cccccc}
				\hline\hline
				&&& \multicolumn{2}{c}{10-shot} &&& &&\multicolumn{2}{c}{20-shot} & \\
				method & $AP$ & $AP_{50}$ & $AP_{75}$ & $AP_{S}$ & $AP_{M}$ & $AP_{L}$  & $AP$  & $AP_{50}$ & $AP_{75}$  & $AP_{S}$  & $AP_{M}$ & $AP_{L}$ \\
				\hline
				Mask R-CNN-ft~\cite{yan2019metarcnn} & 1.9 & 4.7 & 1.3 & 0.2 & 1.4 & 3.2 & 3.7 & 8.5  & 2.9  & 0.3  & 2.5  & 5.8 \\
				Meta R-CNN~\cite{yan2019metarcnn} & 4.4 & 10.6 & 3.3 & {\bf 0.5} & 3.6 & 7.2 & 6.4  & 14.8 & 4.4  & {\bf 0.7}  & 4.9  & 9.3 \\
				CPMask$^*$ (Ours)             & 6.5 & 11.6 & 6.3 & 0.3 & 4.1 & 11.9 & 9.3 & 16.0 & 9.4 & 0.3 & 5.8 & 17.2 \\
				CPMask (Ours)            & {\bf 7.1} & {\bf 12.0} & {\bf 7.2} & 0.3 & {\bf 5.5} & {\bf 12.2} & {\bf 10.3} & {\bf 16.6} & {\bf 10.7} & {\bf 0.7} & {\bf 8.0} & {\bf 17.5} \\
				%\hline
				%Ours (only box)   & 6.8 & 11.5 & 7.0 & 0.2 & 5.2 & 12.0 & 10.0 & 16.2 & 10.6 & {\bf 0.7} & 7.7 & 16.8 \\
				%Ours (no base)   & 4.2 & 9.7 & 3.0 & 0.1 & 3.0 & 8.0 & 6.6 & 13.5 & 5.5 & 0.2 & 5.3 & 12.3 \\
				\hline
		\end{tabular}}}
	\end{center}
	\vspace{-0.1in}
	\caption{Experimental results of few-shot instance segmentation on COCO {\it val} set. The models are trained on the \emph{voc} base categories and fine-tuned on the \emph{non-voc} novel categories with 10/20 instances per category. The evaluation is performed on the held-out \emph{non-voc} novel categories. $^*$ denotes using the Faster R-CNN detector.} %The ``no base" denotes no pre-training on the base categories, namely the model is only trained on the novel categories.} %The "only novel" means the model is only on the novel categories, and the "only box" means the novel categories only have box annotations.} %, %by the realistic protocol, 
	%where good supports are used in fine-tuning or testing.}
	\label{table:few-shot}
	\vspace{-0.2in}
\end{table}

\setlength{\tabcolsep}{3.0pt}
\begin{table}[!t]
	\begin{center}{\small
			\resizebox{0.8\linewidth}{!}{
			\begin{tabular}{cc|c|ccc|ccc}
				\hline\hline
				& method & backbone & $AP$ & $AP_{50}$ & $AP_{75}$ & $AP_{S}$ & $AP_{M}$ & $AP_{L}$ \\ %& speed &GPU \\
				\hline
				%\multicolumn{2}{l|}{{\it two-stage:}}  &&&&&&& \\
				\multirow{4}{*}{Two-stage} & Mask R-CNN~\cite{he2017mask} & ResNet-101 & 35.7 & 58.0 & 37.8 & 15.5 & 38.1 & 52.4 \\ %& & \\
				%& Mask R-CNN$^\star$~\cite{he2017mask} & ResNet101 & 38.3 & {\bf 61.2} & 40.8 & 18.2 & 40.6 & 54.1 \\ %& & \\
				& MaskLab~\cite{chen2018masklab} & ResNet-101 & 37.3 & 59.8 & 39.6 & 19.1 & 40.5 & 50.6 \\
				& MS R-CNN~\cite{huang2019mask} & ResNet-101 & 38.3 & 58.8 & 41.5 & 17.8 & 40.4 & 54.4 \\
				& HTC~\cite{chen2019hybrid} & ResNet-101 & 39.7 & 61.8 & 43.1 & 21.0 & 42.2 & 53.5 \\
				& PANet~\cite{liu2018path} & ResNeXt-101 & {\bf 42.0} & {\bf 65.1} & {\bf 45.7} & {\bf 22.4} & {\bf 44.7} & {\bf 58.1} \\
				\hline
				%\multicolumn{2}{l|}{{\it one-stage:}}  &&&&&&& \\
				\multirow{5}{*}{One-stage}& YOLACT~\cite{bolya2019yolact}       & ResNet-101 & 31.2 & 50.6 & 32.8 & 12.1 & 33.3 & 47.1 \\
				%& PolarMask~\cite{xie2019polarmask}    & ResNet-101 & 32.1 & 53.7 & 33.1 & 14.7 & 33.8 & 45.3 \\
				& TensorMask~\cite{chen2019tensormask}   & ResNet-101 & 37.1 & 59.3 & 39.4 & 17.4 & 39.1 & 51.6 \\
				& ShapeMask~\cite{kuo2019shapemask}      & ResNet-101 & 37.4 & 58.1 & 40.0 & 16.1 & 40.1 & 53.8 \\ %& & \\
				%SOLO       & ResNet101 & 37.8 & 59.5 & 40.4 & 16.4 & 40.6 & 54.2 \\
				& CenterMask~\cite{lee2019centermask}    & ResNet-101 & 38.3 & - & - & 17.7 & 40.8 & {\bf 54.5} \\
				& CPMask (Ours)       & ResNet-101 & {\bf 39.2}    & {\bf 60.8} & {\bf 42.2} & {\bf 22.2} & {\bf 41.8} & 50.1  \\
				%Ours       & ResNext101 &&&&&& \\
				
				\hline
		\end{tabular}}}
	\end{center}
	\vspace{-0.1in}
	\caption{Experimental results of fully supervised instance segmentation on COCO {\it test-dev} set. The mask AP is reported and all entries are single-model results.} %, %by the realistic protocol, 
	%where good supports are used in fine-tuning or testing.}
	\label{table:fully}
	\vspace{-0.35in}
\end{table}

\vspace{-4mm}
\subsection{Fully Supervised Instance Segmentation}
\vspace{-2mm}
In this section we investigate the performance of our model for fully supervised instance segmentation, namely the routine task for instance segmentation.
Table~\ref{table:fully} compares our model with other methods on COCO using COCO \emph{train2017} as train set and \emph{test-dev2017} as test set. The experimental results indicate that our model achieves best performance among one-stage methods, although our method focuses on segmenting novel categories. %Note that our model is also based on one-stage framework.
 Particularly, our model outperforms the best one-stage model CenterMask~\cite{lee2019centermask} by 0.9 AP which is also built on FCOS detection backbone like ours.
These encouraging results proves the effectiveness of model on fully supervised instance segmentation. %The learned commonalities in both shape and appearance is also helpful for supervised learning.

\vspace{-3mm}

\section{Conclusion}
%In this paper, we propose GraphMask to tackle the novel objects segmentation problem. We propose to model general concepts of pixel relation and object boundary to generalize to novel categories. Through our proposed Affinity GCN and Boundary-Aware Network, our model narrows the mask performance gap between novel and base categories where the novel categories are trained with only box annotations. Our method is general and can be applied on different datasets and various annotation scenarios, e.g. fully/partially/few-shot supervised settings.
\vspace{-2mm}
In this paper we present a novel ``\mymodel'' for partially supervised instance segmentation. Our model learns the class-agnostic commonality knowledge that can be generalized from mask-annotated categories to novel categories without mask annotations. Specifically, we design Boundary-Parsing Module to capture shape commonalities by performing supervised learning on boundary estimation. Further, we propose Non-local Affinity-Parsing Module to model pairwise affinities among pixels in intermediate feature maps to learn appearance commonalities across different categories. Benefiting from these two modules, our model outperforms state-of-the-art methods significantly for instance segmentation in both partially-supervised setting and few-shot setting.

 \paragraph{{\bf Acknowledgements}}
 This research is supported in part by the Research Grant Council of the Hong Kong SAR under grant no. 1620818.

\clearpage
\bibliographystyle{splncs04}
\bibliography{main}

\begin{thebibliography}{10}
\providecommand{\url}[1]{\texttt{#1}}
\providecommand{\urlprefix}{URL }
\providecommand{\doi}[1]{https://doi.org/#1}

\bibitem{ahn2019weakly}
Ahn, J., Cho, S., Kwak, S.: Weakly supervised learning of instance segmentation
  with inter-pixel relations. In: CVPR (2019)

\bibitem{arbelaez2009contours}
Arbelaez, P., Maire, M., Fowlkes, C., Malik, J.: From contours to regions: An
  empirical evaluation. In: CVPR (2009)

\bibitem{arbelaez2010contour}
Arbelaez, P., Maire, M., Fowlkes, C., Malik, J.: Contour detection and
  hierarchical image segmentation. IEEE transactions on pattern analysis and
  machine intelligence  \textbf{33}(5),  898--916 (2010)

\bibitem{bertasius2015high}
Bertasius, G., Shi, J., Torresani, L.: High-for-low and low-for-high: Efficient
  boundary detection from deep object features and its applications to
  high-level vision. In: ICCV (2015)

\bibitem{bolya2019yolact}
Bolya, D., Zhou, C., Xiao, F., Lee, Y.J.: Yolact: real-time instance
  segmentation. In: ICCV (2019)

\bibitem{chen2019hybrid}
Chen, K., Pang, J., Wang, J., Xiong, Y., Li, X., Sun, S., Feng, W., Liu, Z.,
  Shi, J., Ouyang, W., et~al.: Hybrid task cascade for instance segmentation.
  In: CVPR (2019)

\bibitem{chen2016semantic}
Chen, L.C., Barron, J.T., Papandreou, G., Murphy, K., Yuille, A.L.: Semantic
  image segmentation with task-specific edge detection using cnns and a
  discriminatively trained domain transform. In: CVPR (2016)

\bibitem{chen2018masklab}
Chen, L.C., Hermans, A., Papandreou, G., Schroff, F., Wang, P., Adam, H.:
  Masklab: Instance segmentation by refining object detection with semantic and
  direction features. In: CVPR (2018)

\bibitem{chen2019tensormask}
Chen, X., Girshick, R., He, K., Doll{\'a}r, P.: Tensormask: A foundation for
  dense object segmentation. In: ICCV (2019)

\bibitem{dai2015convolutional}
Dai, J., He, K., Sun, J.: Convolutional feature masking for joint object and
  stuff segmentation. In: CVPR (2015)

\bibitem{dai2016instance}
Dai, J., He, K., Sun, J.: Instance-aware semantic segmentation via multi-task
  network cascades. In: CVPR (2016)

\bibitem{ding2019boundary}
Ding, H., Jiang, X., Liu, A.Q., Thalmann, N.M., Wang, G.: Boundary-aware
  feature propagation for scene segmentation. In: ICCV (2019)

\bibitem{everingham2010pascal}
Everingham, M., Van~Gool, L., Williams, C.K., Winn, J., Zisserman, A.: The
  pascal visual object classes (voc) challenge. International journal of
  computer vision  \textbf{88}(2),  303--338 (2010)

\bibitem{geiger2012we}
Geiger, A., Lenz, P., Urtasun, R.: Are we ready for autonomous driving? the
  kitti vision benchmark suite. In: CVPR (2012)

\bibitem{georgakis2016multiview}
Georgakis, G., Reza, M.A., Mousavian, A., Le, P.H., Ko{\v{s}}eck{\'a}, J.:
  Multiview rgb-d dataset for object instance detection. In: International
  Conference on 3D Vision (2016)

\bibitem{ghiasi2019fpn}
Ghiasi, G., Lin, T.Y., Le, Q.V.: Nas-fpn: Learning scalable feature pyramid
  architecture for object detection. In: CVPR (2019)

\bibitem{girshick2014rich}
Girshick, R., Donahue, J., Darrell, T., Malik, J.: Rich feature hierarchies for
  accurate object detection and semantic segmentation. In: CVPR (2014)

\bibitem{hariharan2014simultaneous}
Hariharan, B., Arbel{\'a}ez, P., Girshick, R., Malik, J.: Simultaneous
  detection and segmentation. In: ECCV (2014)

\bibitem{hariharan2015hypercolumns}
Hariharan, B., Arbel{\'a}ez, P., Girshick, R., Malik, J.: Hypercolumns for
  object segmentation and fine-grained localization. In: CVPR (2015)

\bibitem{hayder2017boundary}
Hayder, Z., He, X., Salzmann, M.: Boundary-aware instance segmentation. In:
  Proceedings of the IEEE Conference on Computer Vision and Pattern Recognition
  (2017)

\bibitem{he2017mask}
He, K., Gkioxari, G., Doll{\'a}r, P., Girshick, R.: Mask r-cnn. In: ICCV (2017)

\bibitem{he2016deep}
He, K., Zhang, X., Ren, S., Sun, J.: Deep residual learning for image
  recognition. In: CVPR (2016)

\bibitem{hu2018learning}
Hu, R., Doll{\'a}r, P., He, K., Darrell, T., Girshick, R.: Learning to segment
  every thing. In: CVPR (2018)

\bibitem{huang2019mask}
Huang, Z., Huang, L., Gong, Y., Huang, C., Wang, X.: Mask scoring r-cnn. In:
  CVPR (2019)

\bibitem{inoue2018cross}
Inoue, N., Furuta, R., Yamasaki, T., Aizawa, K.: Cross-domain weakly-supervised
  object detection through progressive domain adaptation. In: CVPR (2018)

\bibitem{khoreva2017simple}
Khoreva, A., Benenson, R., Hosang, J., Hein, M., Schiele, B.: Simple does it:
  Weakly supervised instance and semantic segmentation. In: CVPR (2017)

\bibitem{Kong_2018_CVPR}
Kong, S., Fowlkes, C.C.: Recurrent pixel embedding for instance grouping. In:
  CVPR (2018)

\bibitem{kuo2019shapemask}
Kuo, W., Angelova, A., Malik, J., Lin, T.Y.: Shapemask: Learning to segment
  novel objects by refining shape priors. In: ICCV (2019)

\bibitem{lee2019centermask}
Lee, Y., Park, J.: Centermask: Real-time anchor-free instance segmentation. In:
  CVPR (2020)

\bibitem{li2003foreground}
Li, L., Huang, W., Gu, I.Y., Tian, Q.: Foreground object detection from videos
  containing complex background. In: ACM Multimedia (2003)

\bibitem{li2004statistical}
Li, L., Huang, W., Gu, I.Y.H., Tian, Q.: Statistical modeling of complex
  backgrounds for foreground object detection. IEEE Transactions on Image
  Processing  \textbf{13}(11),  1459--1472 (2004)

\bibitem{li2017fully}
Li, Y., Qi, H., Dai, J., Ji, X., Wei, Y.: Fully convolutional instance-aware
  semantic segmentation. In: CVPR (2017)

\bibitem{lin2017feature}
Lin, T.Y., Doll{\'a}r, P., Girshick, R., He, K., Hariharan, B., Belongie, S.:
  Feature pyramid networks for object detection. In: CVPR (2017)

\bibitem{lin2017focal}
Lin, T.Y., Goyal, P., Girshick, R., He, K., Doll{\'a}r, P.: Focal loss for
  dense object detection. In: ICCV (2017)

\bibitem{lin2014microsoft}
Lin, T.Y., Maire, M., Belongie, S., Hays, J., Perona, P., Ramanan, D.,
  Doll{\'a}r, P., Zitnick, C.L.: Microsoft coco: Common objects in context. In:
  ECCV (2014)

\bibitem{liu2018path}
Liu, S., Qi, L., Qin, H., Shi, J., Jia, J.: Path aggregation network for
  instance segmentation. In: CVPR (2018)

\bibitem{luo2017non}
Luo, Z., Mishra, A., Achkar, A., Eichel, J., Li, S., Jodoin, P.M.: Non-local
  deep features for salient object detection. In: CVPR (2017)

\bibitem{mogelmose2012vision}
Mogelmose, A., Trivedi, M.M., Moeslund, T.B.: Vision-based traffic sign
  detection and analysis for intelligent driver assistance systems:
  Perspectives and survey. IEEE Transactions on Intelligent Transportation
  Systems  \textbf{13}(4),  1484--1497 (2012)

\bibitem{pinheiro2015learning}
Pinheiro, P.O., Collobert, R., Doll{\'a}r, P.: Learning to segment object
  candidates. In: NeurIPS (2015)

\bibitem{pinheiro2016learning}
Pinheiro, P.O., Lin, T.Y., Collobert, R., Doll{\'a}r, P.: Learning to refine
  object segments. In: ECCV (2016)

\bibitem{qin2019basnet}
Qin, X., Zhang, Z., Huang, C., Gao, C., Dehghan, M., Jagersand, M.: Basnet:
  Boundary-aware salient object detection. In: CVPR (2019)

\bibitem{remez2018learning}
Remez, T., Huang, J., Brown, M.: Learning to segment via cut-and-paste. In:
  ECCV (2018)

\bibitem{ren2015faster}
Ren, S., He, K., Girshick, R., Sun, J.: Faster r-cnn: Towards real-time object
  detection with region proposal networks. In: NeurIPS (2015)

\bibitem{rother2004grabcut}
Rother, C., Kolmogorov, V., Blake, A.: Grabcut: Interactive foreground
  extraction using iterated graph cuts. ACM transactions on graphics (TOG)
  \textbf{23}(3),  309--314 (2004)

\bibitem{tian2019fcos}
Tian, Z., Shen, C., Chen, H., He, T.: Fcos: Fully convolutional one-stage
  object detection. In: ICCV (2019)

\bibitem{vezhnevets2005growcut}
Vezhnevets, V., Konouchine, V.: Growcut: Interactive multi-label nd image
  segmentation by cellular automata. proc. of Graphicon  \textbf{1},  150--156
  (2005)

\bibitem{wang2019salient}
Wang, W., Zhao, S., Shen, J., Hoi, S.C., Borji, A.: Salient object detection
  with pyramid attention and salient edges. In: CVPR (2019)

\bibitem{wang2018non}
Wang, X., Girshick, R., Gupta, A., He, K.: Non-local neural networks. In: CVPR
  (2018)

\bibitem{wang2019towards}
Wang, X., Cai, Z., Gao, D., Vasconcelos, N.: Towards universal object detection
  by domain attention. In: CVPR (2019)

\bibitem{xia2018dota}
Xia, G.S., Bai, X., Ding, J., Zhu, Z., Belongie, S., Luo, J., Datcu, M.,
  Pelillo, M., Zhang, L.: Dota: A large-scale dataset for object detection in
  aerial images. In: CVPR (2018)

\bibitem{yan2018deep}
Yan, K., Wang, X., Lu, L., Zhang, L., Harrison, A.P., Bagheri, M., Summers,
  R.M.: Deep lesion graphs in the wild: relationship learning and organization
  of significant radiology image findings in a diverse large-scale lesion
  database. In: CVPR (2018)

\bibitem{yan2019metarcnn}
Yan, X., Chen, Z., Xu, A., Wang, X., Liang, X., Lin, L.: Meta r-cnn : Towards
  general solver for instance-level low-shot learning. In: ICCV (2019)

\bibitem{yang2016wider}
Yang, S., Luo, P., Loy, C.C., Tang, X.: Wider face: A face detection benchmark.
  In: CVPR (2016)

\bibitem{yu2018learning}
Yu, C., Wang, J., Peng, C., Gao, C., Yu, G., Sang, N.: Learning a
  discriminative feature network for semantic segmentation. In: CVPR (2018)

\bibitem{pmlr-v97-zhang19f}
Zhang, S., Yan, S., He, X.: {L}atent{GNN}: Learning efficient non-local
  relations for visual recognition. In: ICML (2019)

\bibitem{zhao2019egnet}
Zhao, J.X., Liu, J.J., Fan, D.P., Cao, Y., Yang, J., Cheng, M.M.: Egnet: Edge
  guidance network for salient object detection. In: ICCV (2019)

\bibitem{zhou2018weakly}
Zhou, Y., Zhu, Y., Ye, Q., Qiu, Q., Jiao, J.: Weakly supervised instance
  segmentation using class peak response. In: CVPR (2018)

\end{thebibliography}
\end{document}